\documentclass{article}

\usepackage{fullpage}
\usepackage{graphicx}

\usepackage{url}
\usepackage{graphicx}
\usepackage{subcaption}
\usepackage{amssymb,amsthm, amsmath, dsfont}
\usepackage{lipsum} 
\usepackage{tikz}
\usepackage{pgfplots}
\usepgfplotslibrary{groupplots}
\usepackage{threeparttable}
\usepackage{url}

\usepackage{multirow}
\usepackage{booktabs,array}                 
\usepackage{graphicx}
\usepackage[export]{adjustbox}

\usepackage[linesnumbered,ruled,vlined]{algorithm2e}
\SetKwInput{KwInput}{Input}                
\SetKwInput{KwOutput}{Output}

\newtheorem{assumption}{Assumption}
\newtheorem{proposition}{Proposition}

\newcommand{\R}{{\cal R}}

\newcommand{\reals}{\mathbb{R}}

\DeclareMathOperator*{\jac}{\mathnormal{JAC}}
\newcommand*{\Num}{Number}

\newcommand{\Rand}{RAND}

\begin{document}

\title{Estimating Appearance Models for Image Segmentation via Tensor Factorization}

\author{Jeova F. S. Rocha Neto, \textit{Bowdoin College}}

\maketitle

\begin{abstract}
Image Segmentation is one of the core tasks in Computer Vision, and solving it often depends on modeling the image appearance data via the color distributions of each of its constituent regions. Whereas many segmentation algorithms handle the appearance model dependence using alternation or implicit methods, we propose here a new approach to directly estimate them from the image without prior information on the underlying segmentation. Our method uses local high-order color statistics from the image as an input to a tensor factorization-based estimator for latent variable models. This approach is able to estimate models in multi-region images and automatically output the regions' proportions without prior user interaction, overcoming the drawbacks of a prior attempt to this problem. We also demonstrate the performance of our proposed method in many challenging synthetic and real imaging scenarios and show that it leads to an efficient segmentation algorithm.

\end{abstract}



\section{Introduction}

Image segmentation is a fundamental task in Computer Vision and Image Processing. Given an image with natural regions emerging from the presence of different objects or materials in the depicted scene, the goal of image segmentation is to automatically find these regions. This task finds numerous industrial applications in several fields, such as remote sensing, biomedical imaging, and autonomous vehicles, to name a few \cite{minaee2021image}. Furthermore, segmentation algorithms are consistently used as an intermediate step in higher-level computer vision tasks including object detection and tracking, video processing, and scene understanding. Over the years, numerous algorithms were developed to tackle the segmentation problem, typically using either statistical \cite{ni2009local}, graphical \cite{boykov2001fast, rother2004grabcut, tang2013grabcut}, variational \cite{tsai2001curve, chan2001active, kiechle2018model} or deep-learning \cite{minaee2021image} techniques or combinations of them \cite{xu2007object, xu2017deep, kim2019mumford}. 

Although most of these methods require some level of region appearance description, the algorithms that make use of Markov Random Fields (MRF) \cite{besag1986statistical} explicitly model them in their formulation \cite{barker2000unsupervised, rother2004grabcut}. They approach the segmentation task as an energy minimization problem that encourages the fit of each pixel to the corresponding regional appearance model while promoting spatial coherence of regions \cite{rother2004grabcut, boykov2001fast}. Having these models at the user's disposal, the minimization problem can be framed as a graph-cut problem whose exact solution can be efficiently found for binary segmentation \cite{boykov2001interactive}. In the multiple region setting, the same graph-cut/minimization problem can be approximately solved by iteratively using the 2-regions cut solver \cite{boykov2001fast}.

Unfortunately, appearance model availability is not to be expected in many applications, except for particular instances of biomedical \cite{grady2005multilabel} or remote sensing problems. That led researchers to propose algorithms that either alternate appearance model estimation and segmentation \cite{rother2004grabcut, tang2014pseudo} or that implicitly add the dependency on the appearance to the cut minimization problem \cite{vicente2009joint, tang2013grabcut}. Apart from MRF-based algorithms, others have also attempted to evolve segmentation boundaries to adapt them to local appearance statistics \cite{ni2009local, neto2017level} or to encode the appearance information in pairwise similarities that feed clustering-based approaches \cite{shi2000normalized, felzenszwalb2004efficient}. 

Recently, the authors in \cite{rocha2022direct} proposed a methodology that directly estimates the image's appearance models in images with two regions without explicitly considering the regions they belong to. In their formulation, a set of algebraic expressions involving first and second-order color statistics was derived from simple assumptions about the image formation. From these expressions, efficient appearance estimation algorithms were developed and used in the binary segmentation pipeline to partition challenging image examples. This methodology, however, presented two main limitations: (1) it could only be applied in the context of binary segmentation, without a clear extension to the multi-region setting, and (2) it required a previous guess (or a parameter search) on the region size proportions.

This current work builds on the accomplishments in \cite{rocha2022direct} and proposes an algorithm that can overcome its drawbacks. Inspired by the findings in \cite{anandkumar2014tensor} on the topic of learning latent variable models, we derive an algorithm that effectively estimates the pixel appearance models in multi-region images along with their region site proportions using a tensor factorization-based approach. The presented results also bring evidence that this method can be applied in challenging scenarios where the commonly used segmentation methods fail. In particular, and contrary to the popular practice \cite{mccann2014images, yuan2015factorization, bampis2016projective, kiechle2018model}, we show that one does not need filtering techniques to produce satisfactory texture appearance estimations and segmentations using. This also makes our proposed method faster than comparable methods.

The remainder of this paper is organized as follows. Section II introduces the used notation, defines important variables, and reviews the necessary background for the rest of the paper. Section III presents the assumptions we need to develop our estimation algorithm presented in Section IV. In Section V, we make some implementation remarks that improve our algorithm's practical use. In Section VI, the proposed estimation and segmentation are assessed under various imaging settings and compared to our similar methods. These results are further discussed in the same section. Finally, Section VII concludes the paper and presents further work

\section{Background}
\subsection{Notation}\label{sec:notation}

Let $I : \Omega \rightarrow L$ be an image, where $\Omega$ is the set of
pixel locations and $L$ is a finite set of pixel values, which are scalars in graylevel images and RGB vectors in colored ones. Let $\Delta_{k-1} \triangleq \{x \in \mathbb{R}^k| \sum_i x_i = 1; x_i\geq0, \forall i\}$ be the probability simplex in $k$-dimensions.  We use $I(x)$ to denote the value of a
pixel $x \in \Omega$. An appearance model $\theta \in \Delta_{L-1}$ of a region $\mathcal{R} \subset \Omega$ is a probability distribution that specifies the typical values for the pixels in $\mathcal{R}$\footnote{As pointed out in \cite{rocha2022direct}, appearance models do not necessarily specify a full joint distribution for colors in their regions.}. 

Here we assume that $\Omega$ can be partitioned in $K$ regions $\mathcal{R}_1, \ldots, \mathcal{R}_K$ with $\mathcal{\theta}_1, \ldots, \mathcal{\theta}_K$ as their respective appearance models. We use $w \in \Delta_{K-1}$ to represent the regions' normalized sizes, i.e., for a pixel $x\in \Omega$ selected uniformly at random,  $w_i = \mathbb{P}(x \in \R_i) = |\mathcal{R}_i|/|\Omega|, \forall i$. We also define $S: \Omega \to \{1, \ldots, K\}$ to be a segmentation, i.e., a labeling function that assigns pixels to regions.

Throughout the paper, we'll also make use of the following notation for some high-order statistics required on our proposed algorithm. Let $x, y, z \in \Omega$ be a set of pixels at a distance $r$ from each other\footnote{Since the pixels are in a discrete grid we use the $L_1$ norm to measure the distance between them in practice.}, selected uniformly at random. Let $\alpha \in \reals^L$ be a distribution over $L$ where $\alpha(i) \triangleq \mathbb{P}(I(x) = i)$ is the probability that the pixel $x$ has value $i$.  Let $\beta \in
\reals^{L \times L}$ be a distribution over $L \times L$ where
$\beta(i,j) \triangleq \mathbb{P}(I(x) = i, I(y) = j)$ is the probability that pixel $x$ has value $i$ and pixel $y$ has value $j$. Let $\gamma \in
\reals^{L \times L \times L}$ be a distribution over $L \times L \times L$ where
$\gamma(i, j, k) \triangleq \mathbb{P}(I(x) = i, I(y) = j, I(z) = k)$. 

These distributions can be estimated directly from $I$ by enumerating all pixels in $I$ and all pairs and triplets of pixels that are $r$ pixels apart from each other and accumulating the required occurrences. Define $\widehat{\alpha}$, $\widehat{\beta}$ and $\widehat{\gamma}$ as the estimates for $\alpha$, ${\beta}$ and ${\gamma}$, respectively: 

\begin{equation}\label{eq:alpha}
    \widehat{\alpha}(i) = \frac{1}{|\Omega|}\sum_{x \in \Omega} \mathds{1}_{\{I(x) = i\}},
\end{equation}
\begin{equation}\label{eq:beta}
    \widehat{\beta}(i,j) = \frac{1}{|\mathcal{N}_r|}\sum_{(x,y) \in \mathcal{N}_r} \mathds{1}_{\{I(x) = i\}}) \mathds{1}_{\{I(y) = j\}},
\end{equation}
\begin{equation}\label{eq:gamma}
    \widehat{\gamma}(i,j, k) = \frac{1}{|\mathcal{T}_r|}\sum_{(x,y,z) \in \mathcal{T}_r} \mathds{1}_{\{I(x) = i\}} \mathds{1}_{\{I(y) = j\}}\mathds{1}_{\{I(z) = k\}},
\end{equation}
\noindent where $\mathcal{N}_r = \{ (x,y) \in \Omega^2 \,|\, ||x-y|| = r\}$, $\mathcal{T}_r = \{ (x,y,z) \in \Omega^3 \,|\, ||u-v|| = r, \forall (u,v)\subset (x,y,z)\}$ and $\mathds{1}_{\cdot}$ is the indicator function. Figure 1 gives an example of members of sets $\mathcal{N}_r$ and $\mathcal{T}_r$. Given the grid nature of $I$, we use the $L_1$ to compute distances between pixels. 

\subsection{MRF-based image segmentation}\label{sec:mrf_seg}

Having the appearance models $\mathcal{\theta}_1, \ldots, \mathcal{\theta}_K$, one can use methods based on Markov Random Fields \cite{besag1986statistical} to estimate the underlying segmentation $S$. That is accomplished via the minimization of an energy functional that combines both the appearance models with a boundary regularization term, 
\begin{equation}\label{eq:E_seg}
    E(S |I, \theta_1, \ldots, \theta_K, \lambda) \triangleq
    -\sum_{x \in \Omega} \ln \theta_{S(x)}(I(x)) + \lambda |\partial S|.
\end{equation}
where $|\partial S| = \sum_{(x,y) \in \mathcal{N}_1} \mathds{1}\{S(x)\neq S(y)\}$ and $\lambda$ is a balancing constant. In other words, the above energy accounts for how well the assigned pixels are being fit by the appearance models of the regions they are being assigned to and for how spatially coherent these regions are. Fortunately, solving this energy's minimization problem can be done exactly and efficiently for $K = 2$ and approximately for general $K$ using graph-cut algorithms \cite{boykov2001fast}. 

With this formulation, we can turn the image segmentation task into an appearance estimation problem. The aim of this work is therefore to propose a methodology that estimates these appearance models directly from an image, without considering any prior segmentation. That will be accomplished via the exploration of certain statistical properties of $I$ that arise from some formation assumptions we impose on it.

\subsection{Method of Moments for Latent Variable Models}\label{sec:tensor_approx_explanation}
In this section, we revise the algorithms presented in \cite{anandkumar2014tensor} for estimating latent variables using a Method of Moments strategy. In our current work, we are particularly interested in its proposed solution for the problem of single-topic modeling  with the bag-of-words model for document generation \cite{zhang2010understanding}. This model can be easily converted to our imaging settings, because of the nature of our pixel data, i.e., discrete values corresponding to colors, and to our segmentation setup, as we expect each pixel to belong to only one region in the image.

In the single-topic modeling problem, we have a corpus with documents arising from $K$ different topics and composed of words from a vocabulary with $L$ words. Let $h$ be the latent random variable that corresponds to the topic of a document $D$ of length $\ell$ in the corpus. Let $x_1, x_2, \ldots, x_\ell$ be the words present in $D$. In the single topic model generative process, in order to generate $D$, we first sample $h$ from a discrete distribution $w \in \Delta_{K-1}$, and then we sample $l$ words from the distribution represented by the probability vector $\mu_{h} \in \Delta_{L-1}$. The topic model is therefore fully specified as the set of vectors $(w, \mu_1, \ldots, \mu_K)$ and the goal of the Method of Moments estimator is to estimate these parameters using statistical word moments computed from the documents in our corpus. 

Using one-hot encodings for each word so that $x_1, x_2, \ldots, x_\ell \in \mathbb{R}^L$, one can easily show that $\mathbb{E}(x_i|h) = \mu_h, \forall i \in \{1, \ldots, \ell\}, h \in \{1, \ldots K\}$. Furthermore, one can analytically compute the first, second, and third order moments of our topic model:

\begin{equation}\label{eq:M1}
    M_1 \triangleq \mathbb{E}(x_i) = \sum_{s = 1}^K w_s \mu_s \in \mathbb{R}^{L}
\end{equation}
\begin{equation}\label{eq:M2}
    M_2 \triangleq \mathbb{E}(x_i \otimes x_j) = \sum_{s = 1}^K w_s \mu_s \otimes \mu_s \in \mathbb{R}^{L\times L}
\end{equation}
\begin{equation}\label{eq:M3}
    M_3 \triangleq \mathbb{E}(x_i \otimes x_j  \otimes  x_k) = \sum_{s = 1}^K w_i \mu_s \otimes \mu_s \otimes \mu_s \in \mathbb{R}^{L\times L\times L}
\end{equation}

\noindent where $(i, j, k) \subset \{1, \ldots, K\}$ and  $\otimes$ is the Kronecker product between vectors.

Many algorithms exist in the literature to compute the model parameters from the estimates of $M_1$, $M_2$, and $M_3$ \cite{kolda2009tensor, anandkumar2012method, anandkumar2014tensor}. Our particular focus here will be however on the method used by \cite{kuleshov2015tensor} for its simplicity and speed. Its detailed usage within our appearance model estimation problem is explained in Section \ref{sec:estimation}.

\section{Statistical Model}

\begin{figure}
    \centering
    \includegraphics[trim=5 5 5 5, clip, width=.5\linewidth]{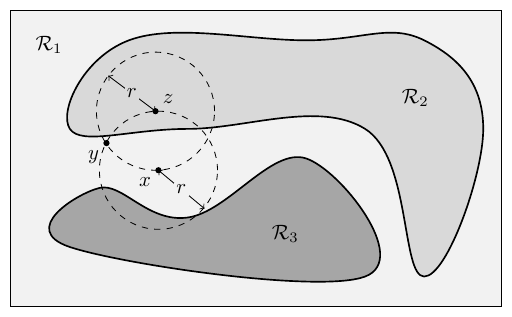}
    \caption{Example of an image with 3 regions. The set $(x, y, z)$ is an exemplar of $\mathcal{T}_r$, while $(x, y)$, $(y, z)$ and $(x, z)$ belong to $\mathcal{N}_r$. }
    \label{fig:example}
\end{figure}

\subsection{Assumptions}\label{sec:assumptions}
Our new approach relies on the same assumptions about the image formation process presented on \cite{rocha2022direct}, restated here for completeness. The first assumption encodes our notion of homogeneity of pixels belonging to a given region. This homogeneity is statistical, in the sense that the marginal distribution of the random process that generated them is the same within that region \cite{ni2009local}.
\begin{assumption}[Statistical Homogeneity]
  \label{as:homogeneity}
  The probability that a pixel $x \in \Omega$ takes a particular value depends
  only on the region the pixel belongs to,
  $$\mathbb{P}(I(x)=i \, | \, x \in \mathcal{R}_s) = \theta_s(i), \forall s \in \{1, \ldots, K\}.$$ 
\end{assumption}

The second assumption captures our understanding that the values of sufficiently far away pixels are independent according to $I$'s generating process. This assumption can be empirically demonstrated in several different imaging scenarios, as shown in \cite{rocha2022direct}.
\begin{assumption}[Independence at a distance]
  \label{as:independence}
  If $x$ and $y$ are two pixels at a distance $r$ from each other,
  \begin{align*}
      \mathbb{P}(I(x)&=i,I(y)=j \, | \, x \in \mathcal{R}_s, y \in \mathcal{R}_t) = 
  \mathbb{P}(I(x) = i \, | \, x \in \mathcal{R}_s) \mathbb{P}(I(y) = j \, | \, y \in \mathcal{R}_t), \forall s \neq t.
  \end{align*}
\end{assumption}

In the present work, we further add a third assumption the boundary of $S$, expected to be short, which entails that the regions delineated by it are spatially coherent. Since this is expected of realistic segmentation, this assumption is usually encoded as a regularization criterion in many image segmentation algorithms \cite{boykov2001interactive, tang2013grabcut}. Here we take it as being a part of the image formation process. 
\begin{assumption}[Short border]
  \label{as:short}
  If $x$ and $y$ are two pixels at a short enough distance $r$ from each other, the probability that they are in different regions is negligible, i.e., 
  $$\mathbb{P}(x \in \mathcal{R}_s, y \in \mathcal{R}_t) = \mathbb{P}(x
\in \mathcal{R}_t, y \in \mathcal{R}_s) \approx 0,\forall s \neq t .$$
\end{assumption}

The choice of the value for $r$ in our algorithms is crucial for the usage of Assumption \ref{as:short} and Figure \ref{fig:example} helps understand why. For small $r$, Assumption \ref{as:short} is clearly satisfied for images with coherent regions, as it is unlikely to randomly select a pair (for the case of $\mathcal{N}_r$) or a triple (for the case of $\mathcal{T}_r$) of pixel locations on or across the region borders. On the other hand, as $r$ increases, this event becomes most frequent, as depicted in Figure \ref{fig:example}. This scenario of larger $r$ is necessary to as, as it allows our methods to be applied to textured images without filtering since it makes Assumption \ref{as:independence} approximately true for them, as shown in \cite{rocha2022direct}. We explore this choice empirically in our experimental section.

\subsection{Moments}

Using Assumptions 1-3 and the variables defined in \ref{sec:notation}, we are ready to state our main proposition.

\begin{proposition}
\label{prop:constraints_multi}
For an image with $K$ regions, under Assumptions 1 and 2, and 3, we have:
\begin{equation}\label{eq:apha_multi}
\alpha = \sum_{s = 1}^K w_s \theta_s.
\end{equation}
\begin{equation}\label{eq:beta_multi}
\beta \approx  \sum_{s = 1}^{K} w_s\theta_s\otimes\theta_s.
\end{equation}
\begin{equation}\label{eq:gamma_multi}
     \gamma \approx  \sum_{s = 1}^{K} w_s\theta_s \otimes\theta_s  \otimes\theta_s.
\end{equation}

\end{proposition}

\begin{proof}

For (\ref{eq:apha_multi}), let $x \in \Omega$ be chosen uniformly at random. Then, we have:
\begin{align*}
    \alpha(i) &= \mathbb{P}(I(x) = i) = \sum_{s = i}^K \mathbb{P}(x \in \R_s) \mathbb{P}(I(x) = i \,|\, x \in \R_s)  = \sum_{s = i}^K w_s \theta_0(i)
\end{align*}
For (\ref{eq:beta_multi}), let $x, y \in \Omega$ with $||x-y||=r$ be chosen uniformly at random. Then, from Assumption \ref{as:short}, we have:
\begin{align*}
    w_s &= \mathbb{P}(x \in \R_s) = \sum_{t = 1}^{K}\mathbb{P}(x \in \R_s, y \in \R_t) = \mathbb{P}(x \in \R_s, y \in \R_s) = \mathbb{P}(x, y \in \R_s), \forall s \in \{1, \ldots, K\}
\end{align*}
\noindent Now, using Assumption \ref{as:short}, we have:
\begin{align*}
    \beta(i,j) &= \mathbb{P}(I(x) = i, I(y) = j) \\
    & =  \sum_{s = 1}^{K}\mathbb{P}(x, y \in \R_s) \mathbb{P}(I(x) = i, I(y) = j | x,y \in \R_s)\\
    & \approx    \sum_{s = 1}^{K} w_s \mathbb{P}(I(x) = i|x \in \R_s)\mathbb{P}(I(y) = j|y \in \R_s)  \\
    & = \sum_{s = 1}^{K} w_s \theta(i)\theta(j),
\end{align*}

From Assumption \ref{as:short}, we have that 
\begin{align*}
    w_s &= \mathbb{P}(x \in \R_s) = \sum_{t = 1}^{K}\sum_{u = 1}^{K}\mathbb{P}(x \in \R_s, y \in \R_t, z \in \R_u) = \sum_{t = 1}^{K}\sum_{u = 1}^{K} \mathbb{P}(x \in \R_s, y \in \R_t)\mathbb{P}(z \in \R_u|x \in \R_s, y \in \R_t)
    \\&= \sum_{u = 1}^{K} \mathbb{P}(x \in \R_s, y \in \R_s)\mathbb{P}(z \in \R_u|x \in \R_s, y \in \R_s)
    = \sum_{u = 1}^{K} \mathbb{P}(x \in \R_s, z \in \R_u)\mathbb{P}(y \in \R_s|x \in \R_s, z \in \R_u)
    \\&=  \mathbb{P}(x \in \R_s, z \in \R_s)\mathbb{P}(y \in \R_s|x \in \R_s, z \in \R_s)
    =  \mathbb{P}(x \in \R_s, y \in \R_s, z \in \R_s) 
    \\&= \mathbb{P}(x, y, z \in \R_s), \forall s \in \{1, \ldots, K\}
\end{align*}

Now, for (\ref{eq:gamma_multi}),
\begin{align*}
    \gamma(i,j,k) &= \mathbb{P}(I(x) = i, I(y) = j, I(z) = k) 
     \\&= \sum_{s = 1}^{K}\mathbb{P}(x, y, z \in \R_s)  \mathbb{P}(I(x) = i, I(y) = j, I(z) = k | x,y, z \in \R_s) 
    \\&\approx  \sum_{s = 1}^{K} w_s \mathbb{P}(I(x) = i|x \in \R_s)   \mathbb{P}(I(y) = j|y \in \R_s) \mathbb{P}(I(z) = k | z \in \R_s) 
    \\&= \sum_{s = 1}^{K} w_s \theta(i)\theta(j)\theta(k),
\end{align*}
where the third equality is due to the independence assumption from Assumption \ref{as:independence}.
\end{proof}

\section{Appearance Estimation}\label{sec:estimation}

In Eqs. \ref{eq:apha_multi}-\ref{eq:gamma_multi} there is a clear application of the more general estimation setting proposed presented in Section \ref{sec:tensor_approx_explanation}, where the model distributions $\{\mu_h\}$ and topic proportions correspond to our desired appearance models $\{\theta_k\}$ and region size proportions, respectively. This correspondence entails also the correspondence between $\alpha$, $\beta$, and $\gamma$, under the requirements of Proposition 1, and the moments depicted in Eqs. \ref{eq:M1}-\ref{eq:M3}. These connections lead to the interpretation of our appearance model estimation task as a single-topic modeling problem, having different discrete colors corresponding to words from the vocabulary and image regions corresponding to topics.

Following, we derive our estimator based on the algorithm initially proposed in \cite{kuleshov2015tensor} and refined in \cite{ruffini2017hierarchical}. Let $\Theta = [\theta_1 | \theta_2 |\cdots | \theta_K] \in \mathbb{R}^{L \times K}$ and $W = \operatorname{diag}(w) \in \mathbb{R}^{K \times K}$ as the diagonal matrix whose elements are the values in $w$. Now, from Eqs. \ref{eq:beta_multi} and \ref{eq:gamma_multi}, we can write ${\beta}$ and the $s$-th slice of ${\gamma}$, denoted as ${\gamma}(\cdot, \cdot, s)$, as:
\begin{equation}
    \mathbf{ {\beta}} =  \Theta W \Theta^{\top}, 
\end{equation}
\begin{equation}
    {\gamma}(\cdot, \cdot, s) = \Theta W^{\frac{1}{2}} \operatorname{diag}(\Theta(s, \cdot))W^{\frac{1}{2}}\Theta^{\top},
\end{equation}
where $ \Theta(s, \cdot) = [\theta_1(s), \ldots, \theta_K(s)]$ is the $s$-th row of $\Theta$. From the above expression, we note that $ {\beta}$ is positive semidefinite of rank $K$, and therefore we can compute its SVD as $ {\beta} = U\Lambda U^{\top}$, truncated at the $K$-th singular vector.  Now, defining $M = U\Lambda^{\frac{1}{2}}$ leads to
\begin{equation}\label{eq:mo}
   \Theta W^{\frac{1}{2}} = MO,
\end{equation}
for an unique orthonormal matrix $O$. Once $O$ is estimated, we can then estimate $\Theta$.

The authors in \cite{anandkumar2012method} use the pseudo inverse of $M$, here defined as $M^{\dagger}$, as a whitening step on the slices of $ {\gamma}$. In their algorithm, this step was used to transform $\gamma$ into a symmetric orthogonally decomposable tensor whose high-order singular vectors could efficiently be estimated via a tensorial algorithm analogous to the power method in matrices. The slices of this whitened tensor $ {\gamma}^{\text{white}}$ now present the following property:
\begin{equation}
     {\gamma}^{\text{white}}(\cdot, \cdot, s)  = M^{\dagger}  {\gamma}(\cdot, \cdot, s) (M^{\dagger})^{\top} = O  \operatorname{diag}(\Theta(s, \cdot)) O^{\top}, \forall s \in \{1, \ldots, K\}.
\end{equation}
The above equation tells us that there must be a unique orthonormal matrix $O$ that diagonalizes all the slices of $ {\gamma}^{\text{white}}$ simultaneously. For $K = 2$, an approximation of $O$ can be found in constant time by the method proposed by \cite{ruffini2017hierarchical} and, for $K \geq 2$, the algorithm proposed in \cite{cardoso1996jacobi} approximately solves it in polynomial time. 

With an approximate $\widehat{O}$, an approximate solution to $\Theta$ can be found as $ \widehat{\Theta} = M\widehat{O}$, followed by the projection of its columns to $\Delta_{L-1}$. Note this  projection step frees us from an earlier necessity for an estimate of $W$ (ref. Eq \ref{eq:mo}) and also assures that the estimated models are distributions. Finally, noticing that $ {\alpha} = \Theta w$ from Eq. \ref{eq:apha_multi}, an approximation of $w$ can be computed as $\widehat{w} = \widehat{\Theta}^{\dagger} {\alpha}$ followed by its projection to  $\Delta_{K-1}$. 

\section{Implementation remarks}
Implementing the above algorithm for imaging purposes can suffer from three computational hindrances for its practical use. In the following, we describe each issue and their proposed solutions.

\subsection{Estimation of $\beta$ and $\gamma$ }
The enumeration of all pixels triplets in $\mathcal{T}_r$ for the estimation of $\gamma$ according to Eq. \ref{eq:gamma} can be computationally expensive for a large, high-resolution image. In practice, we only consider, for a given pixel $x$ at $(x_1, x_2)$ location in $\Omega$, all triples involving the $x$, the pixels at $(x_1 \pm r, x_2)$ and at $(x_1, x_2 \pm r)$. Despite overlooking most triples in $\mathcal{T}_r$, this approach yields a considerable gain in estimation speed and did not hurt our estimations in practice.

Furthermore, Eqs. \ref{eq:alpha}-\ref{eq:gamma} represent a different enumeration strategy compared to what is described in \cite{rocha2022direct}. There, the authors proposed to compute $\alpha$ from the marginalization of $\beta$, which makes a better use of the pixels near the image borders for estimation. This process could also be implemented in our application by marginalizing $\gamma$ to have estimates for both $\alpha$ and $\beta$. However, because of the sparser nature of $\gamma$'s estimation, described above, and the algorithmic consideration laid in Section \ref{seg:dim_gamma}, we decided for the separate enumerations of individual, pairs and triplets of pixels.

Finally, for very large images, one can also sample pairs and triples of pixels from $I$ as an alternative approach to enumerating them all to estimate $\beta$ and $\gamma$.

\subsection{Dimension of $\gamma$} \label{seg:dim_gamma}
For relatively large $L$, the dimensionality of $\gamma \in \mathbb{R}^{L\times L \times L}$ can impose a considerable memory burden on this algorithm in practice. The problem can be avoided by noticing that the estimation procedure outlined in Section \ref{sec:estimation} does not require $\gamma$ in its full size, but only its slices separately, so it can proceed with the simultaneous diagonalization step. Furthermore, the method also does not require $\gamma$ to be normalized. Taking advantage of that fact, we only need to estimate certain slices of $\gamma$ and we can skip the normalization step, which would require the tensor in its full size.

In practice, we consider the slices corresponding to the $\ell < L$ colors found most frequently in $I$ and disregard the slices concerning the other colors, making $\gamma$ effectively be of dimension $L \times L \times \ell$. In our experimental section, we empirically show that the absence of these slices for the diagonalization procedure does not impose a large loss on our method's estimation performance. 

The overall estimation approach is shown in Algorithm \ref{alg:estimation_moments}. 

\subsection{Color images}\label{sec:color_imgs}
Our estimation algorithm only depends on $L$, the number of unique color values in $I$. Although that number is relatively small for grayscale images, upper bounded by 256, it can be substantially large for RBG images. To handle this issue, we proceed as in \cite{rocha2022direct} and added a pre-processing step to our algorithm to reduce the total number of colors in RGB images to a smaller number of quantized values. In this work, we use a hierarchical $K$-means approach that recursively partitions the color space in two until the total number of partitions reaches a predefined amount $N$. 

\subsection{Algorithm's Overview}

Throughout this paper, we shall name our proposed appearance model estimation, which comprises the estimation of $\alpha$, $\beta$, and $\gamma$ along with the subsequent $\theta$ and $w$ estimations, of \textbf{T}ensorial \textbf{E}stimator of \textbf{A}ppearance \textbf{M}odels (TEAM), summarized in Algorithm \ref{alg:estimation_models}. When coupled with the graph-cut based segmentation step (see Section \ref{sec:segmentaion}), it shall be called TEAMSEG\footnote{TEAMSEG's code can be found at \url{github.com/jeovafarias/teamseg}.}.

\begin{algorithm}[!ht]
\caption{Moment Estimation Algorithm}\label{alg:estimation_moments}
\DontPrintSemicolon
  
    \KwInput{Image: $I$, Radius: $r$, \Num\ of Slices: $\ell$}
    \KwOutput{Color moment estimates: $\widehat{\alpha}$, $\widehat{\beta}$, $\widehat{\gamma}$}
    
    Compute $\widehat{\alpha}$ as the normalized histogram of $I$
    
    Initialize $\widehat{\beta} \in \mathbb{R}^{L \times L}$ and $\widehat{\gamma} \in \mathbb{R}^{L \times L \times L}$ with zeros.
    
    Define $F$ as the set of $\ell$ most frequent colors in $I$.
    
    \ForEach{$x \in \Omega \subset \mathbb{Z}^2$}{
        Define $v_1 = I(x)$
        
        \ForEach{$y \in \Omega$ \textbf{and} $\lVert x - y \rVert = r$}{
            Increment $\widehat{\beta}(v_1, I(y))$ and $\widehat{\beta}(I(y), v_1)$
        }
        Define $v_2 = I(x + (0, r))$ and $v_3 = I(x + (r, 0))$
        
        \If{$v_1 \in F$}{ 
            Increment $\widehat{\gamma}(v_2, v_3, v_1)$ and $\widehat{\gamma}(v_3, v_2, v_1)$
        }
        
        \If{$v_2 \in F$}{ 
            Increment $\widehat{\gamma}(v_1, v_3, v_2)$ and $\widehat{\gamma}(v_3, v_1, v_3)$
        }
        
        \If{$v_3 \in F$}{ 
            Increment $\widehat{\gamma}(v_1, v_2, v_3)$ and $\widehat{\gamma}(v_2, v_1, v_3)$
        }
    }
    
    Normalize $\widehat{\beta}$ to be a distribution in $\mathbb{R}^{L\times L}$

\end{algorithm}

\begin{algorithm}[!ht]
\caption{TEAM Algorithm}\label{alg:estimation_models}
\DontPrintSemicolon
  
  \KwInput{Image: $I$, Radius: $r$, \Num\ of Slices: $\ell$, \Num\ of Regions: $K$}
  \KwOutput{Estimates for the appearance models and region size proportions: $\widehat{w}, \widehat{\theta_1}, \ldots, \widehat{\theta_K}$}
    If $L$ is large, partition the color space using method from Section \ref{sec:color_imgs}. 
    
    Compute $\widehat{\alpha}$, $\widehat{\beta}$, $\widehat{\gamma}$ from  Algorithm \ref{alg:estimation_moments} using $I$, $r$ and $\ell$. 
  
    Compute the SVD of $\widehat{\beta}$ truncated at the $K$-th singular vector to obtain $U \in \mathbb{R}^{L \times K}$ and $\Lambda \in \mathbb{R}^{K\times K}$.
    
    Define $M = U\Lambda^{\frac{1}{2}}$.
    
    Compute the pseudo-inverse of $M$: $M^{\dagger}$.

    \For{$r=1$ \KwTo $\ell$}
    {
        $\gamma^{\text{white}}(\cdot, \cdot, r) = M^{\dagger} \gamma(:, :, r) (M^{\dagger})^{\top}$
    }
    
    \eIf{K = 2}{
        Find $O$ that diagonilizes $\gamma^{\text{white}}$ using alg. in \cite{ruffini2017hierarchical}.
    }{
        Find $O$ that diagonilizes $\gamma^{\text{white}}$ using alg. in \cite{cardoso1996jacobi}.
    }
    
    Project the columns of $MO$ to $\Delta_{L-1}$ to obtain $\widehat{\Theta} = [\widehat{\theta}_1, \ldots, \widehat{\theta}_K]$
    
    Project $\widehat{\Theta}^{\dagger} {\alpha}$ to $\Delta_{K-1}$  to obtain $\widehat{w}$.
\end{algorithm}

\section{Experiments and Discussion}

\subsection{Segmentation Algorithm}\label{sec:segmentaion}
As explained in Section \ref{sec:mrf_seg}, we can obtain a segmentation of $I$ using the estimated appearance modes via an MRF-based energy minimization problem (see Eq. \ref{eq:E_seg}). For the case when $K = 2$, the authors in \cite{boykov2001interactive} describe how to derive a graph consisting of a grid connecting the image pixels and their immediate neighbors, and two extra nodes, $s$ and $t$, representing each image region to be detected. There are edges connecting each grid node to $s$ and to $t$, such that their weight corresponds to the negative loglikelihoods of each pixel color being assigned to a region, according to the available appearance models. The grid weights are set to $\lambda$ to correspond to the formulation in Eq. \ref{eq:E_seg}. With this graph constructed, one can minimize the energy in Eq. \ref{eq:E_seg} by performing an $st$-cut on it. 

For the multi-region case, the authors in \cite{boykov2001fast} showed that an approximate solution to Eq. \ref{eq:E_seg} minimization can be accomplished by a series of large "moves", where a similar graph as before is computed on which a $st$-cut is performed. Of the proposed two types of moves, for simplicity, we make only use here of $\alpha\beta$-swaps moves, where all pairs of labels are optimized iteratively until convergence. 


\subsection{Evaluation Metrics, Methods Compared and Hardware}

Following \cite{rocha2022direct} we use the Bhattacharyya distance to evaluate the quality of the appearance models we estimate having the models defined by a ground truth segmentation as a standard. For $p, q \in \Delta_{L-1}$, the Bhattacharyya distance $d_{B}(p, q)$ is defined as:
\begin{equation}
    d_{B}(p, q) \triangleq  -\ln\left( \sum_{i \in L} \sqrt{p(i)q(i)}\right) \in [0, +\infty).
\end{equation}

\noindent Let ${\theta}_1, \ldots, {\theta}_K$ be the ground-truth appearances and $\widehat{\theta}_1, \ldots, \widehat{\theta}_K$ our estimates for them. Since the region and appearance labels are exchangeable, we have to consider all possible matchings between the estimated and ground-truth models in our evaluation. Let $Perm(K)$ be the set of all permutations of the sequence $\{1, \ldots, K\}$. Our appearance estimation evaluation measure $D_{B}$ is then defined as:
\begin{equation}\label{eq:multi_db}
    D_{B} \triangleq \min_{\pi \in Perm(K)}\left(\frac{1}{K} \sum_{s = 1}^K d_{B}\left(\theta_s, \widehat{\theta}_{\pi(s)}\right)\right).
\end{equation}

\begin{figure}
\centering
    \input{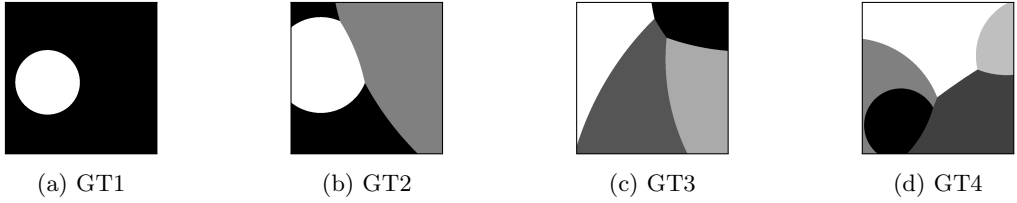}
    \caption{Ground-truth segmentation masks used in our synthetic experiments. Image frames are not part of the original images.}
    \label{fig:gts}
\end{figure}

\begin{figure}
\centering
    \input{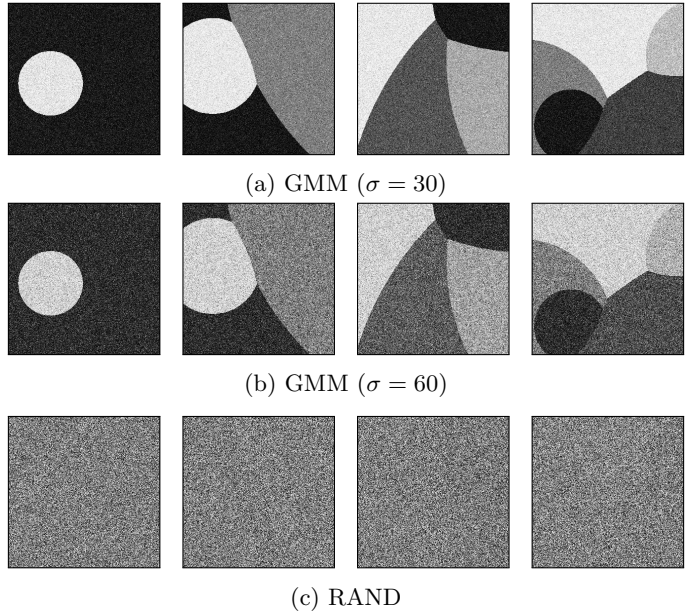}
    \caption{Example of synthetic images generated from each generation process and for each segmentation mask used in our experiments. Image frames are not part of the original images.}
    \label{fig:example_synth_imgs}
\end{figure}

As our method also estimates region proportions, we use $d_{B}(w, \widehat{w})$ to compare an estimated region proportion vector $\hat{w}$ to its ground-truth counterpart $w$.

To evaluate the accuracy of our segmentation results, we consider the overlap between two regions $\mathcal{R}_s,\mathcal{R}_t \subseteq \Omega$ in different segmentations using the Jaccard index,
\begin{equation}\label{eq:jaccard}
    j(\mathcal{R}_s, \mathcal{R}_t) \triangleq |\mathcal{R}_s \cap \mathcal{R}_t|/|\mathcal{R}_s\cup \mathcal{R}_t| \in [0, 1],
\end{equation}
\noindent where larger indexes correspond to larger overlaps. Let $\R_1, \ldots, \R_K$ and $\widehat{\R}_1, \ldots, \widehat{\R}_K$ are the ground truth and estimated segmentations, respectively. We use the following measure to evaluate our segmentation performance in the multi-region setting:
\begin{equation}
\jac = \max_{\pi \in Perm(K)}\left(\frac{1}{K}\sum_{s = 1}^K \left(j(\R_s,\widehat{\R}_{\pi(s)}\right)\right).
\end{equation}

As a baseline comparative method, we consider the Expectation-Maximization algorithm \cite{dempster1977maximum} for estimating the parameters of a Gaussian Mixture Model (here called EM-GMM) on the image pixel values. After convergence, we discretize the estimated Gaussians by considering the values of those densities at the discrete intervals in $0, 1, \ldots, 256$ as vector and projecting it to $\Delta_{255}$.  Here we also compare our proposed estimator to both algebraic (ALGB) and spectral (SPEC) estimators proposed in \cite{rocha2022direct} for the experiments with $K = 2$ and to the estimation/segmentation alternation approach (ALT) described in that same work. 

On the segmentation side, we consider two of the comparative methods described in \cite{rocha2022direct}, Images as Occlusions of Textures (ORTSEG) \cite{mccann2014images} and Factorization Based Segmentation (FBS) \cite{yuan2015factorization}. Adding to those, we also consider Graph Projective Non-Negative Matrix Factorization (GRPNMF) \cite{bampis2016projective}. The methods were chosen as they (1) also assume either explicitly or implicitly that regions have homogeneous appearance, (2) have a window size parameter that functions similarly to $r$ in our methods, (3) can tackle multi-region segmentation problems, and (4) have official Matlab implementation available. Each method was tuned to achieve its best performances. We also compare the model estimation image segmentation performances of ALGB and SPEC when coupled with the graph-cut algorithm described in Section \ref{sec:segmentaion} and of ALT.

The following experiments were performed using code written in MATLAB and run on an Intel\textregistered \, Core\textsuperscript{TM} i5-6200U CPU $@$ 2.30GHz with 8GB of RAM.

\subsection{Synthetic Experiment Setup}
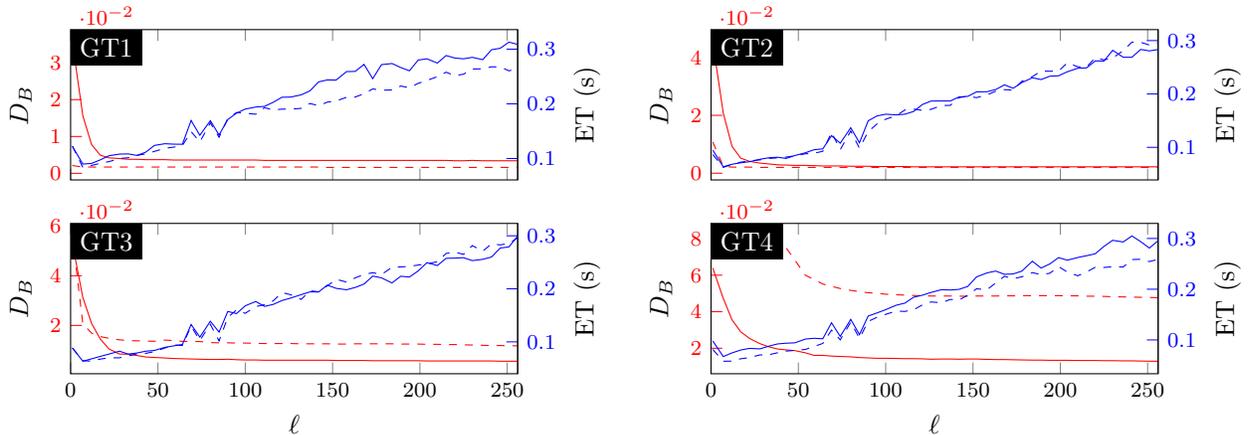
\begin{figure}

    \newcommand\Width{.36\linewidth}

\begin{tikzpicture}
        \begin{axis}[
            scale only axis,
            height=2cm,
            width=\Width,
            ytick pos=left,
            ylabel={$D_{B}$},
            yticklabel style = {font=\footnotesize, color=red},
            xticklabel style = {font=\footnotesize},
            every y tick/.style={red},
            ylabel near ticks,
            xlabel near ticks,
            xmin=0,xmax=256,
            xtick = {},
            xticklabels = {},
            ]
        \addplot[mark=none, red] table[x index=0,y index=1] {vary_slices/vary_slices_gt1_0.dat};\label{line:slices_nogmm_db}
        \addplot[mark=none, red, dashed] table[x index=0,y index=1] {vary_slices/vary_slices_gt1_1.dat};\label{line:slices_gmm_db}
        
        \draw (rel axis cs:0,.88) node[right,text=white, fill=black,align=left] {GT1};
        \end{axis}
        
        \begin{axis}[
            scale only axis,
            height=2cm,
            width=\Width,
            axis y line*=right,
            axis x line=none,   
            ylabel={ET (s)},
            yticklabel style = {font=\footnotesize, color=blue},
            xticklabel style = {font=\footnotesize},
            every y tick/.style={blue},
            ylabel near ticks,
            xlabel near ticks,
            xmin=0,xmax=256,   
            ]
        \addplot[mark=none, blue] table[x index=0,y index=2] {vary_slices/vary_slices_gt1_0.dat};\label{line:slices_nogmm_et}
        \addplot[mark=none, blue, dashed] table[x index=0,y index=2] {vary_slices/vary_slices_gt1_1.dat};\label{line:slices_gmm_et}
        
        \end{axis}
    \end{tikzpicture}%
    \hspace{10pt}
    \begin{tikzpicture}
        \begin{axis}[
            scale only axis,
            height=2cm,
            width=\Width,
            ytick pos=left,
            ylabel={$D_{B}$},
            yticklabel style = {font=\footnotesize, color=red},
            every y tick/.style={red},
            ylabel near ticks,
            xlabel near ticks,
            xmin=0,xmax=256,
            xtick = {},
            xticklabels = {},
            ]
        \addplot[mark=none, red] table[x index=0,y index=1] {vary_slices/vary_slices_gt2_0.dat};
        \addplot[mark=none, red, dashed] table[x index=0,y index=1] {vary_slices/vary_slices_gt2_1.dat};
        
        \draw (rel axis cs:0,.88) node[right,text=white, fill=black,align=left] {GT2};
        \end{axis}
        
        \begin{axis}[
            scale only axis,
            height=2cm,
            width=\Width,
            axis y line*=right,
            axis x line=none,   
            ylabel={ET (s)},
            yticklabel style = {font=\footnotesize, color=blue},
            xticklabel style = {font=\footnotesize},
            every y tick/.style={blue},
            ylabel near ticks,
            xlabel near ticks,
            xmin=0,xmax=256,   
            ]
        \addplot[mark=none, blue] table[x index=0,y index=2] {vary_slices/vary_slices_gt2_0.dat};
        \addplot[mark=none, blue, dashed] table[x index=0,y index=2] {vary_slices/vary_slices_gt2_1.dat};
        \end{axis}
    \end{tikzpicture}
    \begin{tikzpicture}
        \begin{axis}[
            scale only axis,
            height=2cm,
            width=\Width,
            ytick pos=left,
            xlabel={$\ell$},
            ylabel={$D_{B}$},
            yticklabel style = {font=\footnotesize, color=red},
            xticklabel style = {font=\footnotesize},
            every y tick/.style={red},
            ylabel near ticks,
            xlabel near ticks,
            xmin=0,xmax=256,
            scaled y ticks=true,
            ]
        \addplot[mark=none, red] table[x index=0,y index=1] {vary_slices/vary_slices_gt3_0.dat};
        \addplot[mark=none, red, dashed] table[x index=0,y index=1] {vary_slices/vary_slices_gt3_1.dat};
        \draw (rel axis cs:0,.88) node[right,text=white, fill=black,align=left] {GT3};
        \end{axis}
        
        \begin{axis}[
            scale only axis,
            height=2cm,
            width=\Width,
            axis y line*=right,
            axis x line=none,   
            ylabel={ET (s)},
            yticklabel style = {font=\footnotesize, color=blue},
            xticklabel style = {font=\footnotesize},
            every y tick/.style={blue},
            ylabel near ticks,
            xlabel near ticks,
            xmin=0,xmax=256,   
            ]
        \addplot[mark=none, blue] table[x index=0,y index=2] {vary_slices/vary_slices_gt3_0.dat};
        \addplot[mark=none, blue, dashed] table[x index=0,y index=2] {vary_slices/vary_slices_gt3_1.dat};
        \end{axis}
    \end{tikzpicture}%
    \hspace{10pt}
    \begin{tikzpicture}
        \begin{axis}[
            scale only axis,
            height=2cm,
            width=\Width,
            ytick pos=left,
            xlabel={$\ell$},
            ylabel={$D_{B}$},
            yticklabel style = {font=\footnotesize, color=red},
            xticklabel style = {font=\footnotesize},
            every y tick/.style={red},
            ylabel near ticks,
            xlabel near ticks,
            xmin=0,xmax=256,
            scaled y ticks=true,
            ]
        \addplot[mark=none, red] table[x index=0,y index=1] {vary_slices/vary_slices_gt4_0.dat};
        \addplot[mark=none, red, dashed] table[x index=0,y index=1] {vary_slices/vary_slices_gt4_1.dat};
        
        \draw (rel axis cs:0,.88) node[right,text=white, fill=black,align=left] {GT4};

        \end{axis}
        
        \begin{axis}[
            scale only axis,
            height=2cm,
            width=\Width,
            axis y line*=right,
            axis x line=none,   
            ylabel={ET (s)},
            yticklabel style = {font=\footnotesize, color=blue},
            xticklabel style = {font=\footnotesize},
            every y tick/.style={blue},
            ylabel near ticks,
            xlabel near ticks,
            xmin=0,xmax=256,
            ]
        \addplot[mark=none, blue] table[x index=0,y index=2] {vary_slices/vary_slices_gt4_0.dat};
        \addplot[mark=none, blue, dashed] table[x index=0,y index=2] {vary_slices/vary_slices_gt4_1.dat};
        \end{axis}
    \end{tikzpicture}
    \vspace{-15pt}
    \caption{Varying the number of slices $\ell$ of $\gamma$ when estimating the appearance models of images from the four proposed ground-truth masks. The timing (\ref{line:slices_gmm_et}) and performance (\ref{line:slices_gmm_db}) of our method on GMM-generated images ($\sigma = 60$) are shown in dashed lines, whereas the solid lines (\ref{line:slices_nogmm_et} and \ref{line:slices_nogmm_db}) are for images generated from \Rand. The lines show the average result of the estimations on 50 synthetic images of size $300 \times 300$ and with $L = 256$ colors.} 
    \label{fig:vary_slices}
\end{figure}

\begin{figure}
    \newcommand\Width{.36\linewidth}

    \begin{tikzpicture}
        \begin{axis}[
            scale only axis,
            height=2cm,
            width=\Width,
            ytick pos=left,
            ylabel={$d_{B}$},
            yticklabel style = {font=\footnotesize, color=red},
            xticklabel style = {font=\footnotesize},
            every y tick/.style={red},
            ylabel near ticks,
            xlabel near ticks,
            xmin=0,xmax=128,
            ymin=0,ymax=.4,
            xtick = {},
            xticklabels = {},
            ]
        \addplot[mark=none, red] table[x index=0,y index=1] {vary_noise/vary_noise_gt1.dat}; \label{line:mom_db}
        \addplot[mark=none, red, dashed] table[x index=0,y index=1] {vary_noise/vary_noise_gt1_em.dat};\label{line:em_db}

        \draw (rel axis cs:0,.88) node[right,text=white, fill=black,align=left] {GT1};
        \end{axis}
        
        \begin{axis}[
            scale only axis,
            height=2cm,
            width=\Width,
            axis y line*=right,
            axis x line=none,   
            ylabel={$\jac$},
            yticklabel style = {font=\footnotesize, color=blue},
            xticklabel style = {font=\footnotesize},
            every y tick/.style={blue},
            ylabel near ticks,
            xlabel near ticks,
            xmin=0,xmax=128,  
            ymin=0.2,ymax=1, 
            ]
        \addplot[mark=none, blue] table[x index=0,y index=2] {vary_noise/vary_noise_gt1.dat};\label{line:mom_jac}
        \addplot[mark=none, blue, dashed] table[x index=0,y index=2] {vary_noise/vary_noise_gt1_em.dat};\label{line:em_jac}
        
        \end{axis}
    \end{tikzpicture}%
    \begin{tikzpicture}
        \begin{axis}[
            scale only axis,
            height=2cm,
            width=\Width,
            ytick pos=left,
            ylabel={$d_{B}$},
            yticklabel style = {font=\footnotesize, color=red},
            xticklabel style = {font=\footnotesize},
            every y tick/.style={red},
            ylabel near ticks,
            xlabel near ticks,
            xmin=0,xmax=128,
            ymin=0,ymax=.4,
            xtick = {},
            xticklabels = {},
            ]
        \addplot[mark=none, red] table[x index=0,y index=1] {vary_noise/vary_noise_gt2.dat};
        \addplot[mark=none, red, dashed] table[x index=0,y index=1] {vary_noise/vary_noise_gt2_em.dat};

        \draw (rel axis cs:0,.88) node[right,text=white, fill=black,align=left] {GT2};
        \end{axis}
        
        \begin{axis}[
            scale only axis,
            height=2cm,
            width=\Width,
            axis y line*=right,
            axis x line=none,   
            ylabel={$\jac$},
            yticklabel style = {font=\footnotesize, color=blue},
            xticklabel style = {font=\footnotesize},
            every y tick/.style={blue},
            ylabel near ticks,
            xlabel near ticks,
            xmin=0,xmax=128,  
            ymin=0.2,ymax=1, 
            ]
        \addplot[mark=none, blue] table[x index=0,y index=2] {vary_noise/vary_noise_gt2.dat};
        \addplot[mark=none, blue, dashed] table[x index=0,y index=2] {vary_noise/vary_noise_gt2_em.dat};

        \end{axis}
    \end{tikzpicture}
    
    \begin{tikzpicture}
        \begin{axis}[
            scale only axis,
            height=2cm,
            width=\Width,
            ytick pos=left,
            ylabel={$d_{B}$},
            xlabel={$\sigma$},
            yticklabel style = {font=\footnotesize, color=red},
            xticklabel style = {font=\footnotesize},
            every y tick/.style={red},
            ylabel near ticks,
            xlabel near ticks,
            xmin=0,xmax=128,
            ymin=0,ymax=.4,
            ]
        \addplot[mark=none, red] table[x index=0,y index=1] {vary_noise/vary_noise_gt3.dat};
        \addplot[mark=none, red, dashed] table[x index=0,y index=1] {vary_noise/vary_noise_gt3_em.dat};

        \draw (rel axis cs:0,.88) node[right,text=white, fill=black,align=left] {GT3};
        \end{axis}
        
        \begin{axis}[
            scale only axis,
            height=2cm,
            width=\Width,
            axis y line*=right,
            axis x line=none,   
            ylabel={$\jac$},
            yticklabel style = {font=\footnotesize, color=blue},
            xticklabel style = {font=\footnotesize},
            every y tick/.style={blue},
            ylabel near ticks,
            xlabel near ticks,
            xmin=0,xmax=128,  
            ymin=0.2,ymax=1, 
            ]
        \addplot[mark=none, blue] table[x index=0,y index=2] {vary_noise/vary_noise_gt3.dat};
        \addplot[mark=none, blue, dashed] table[x index=0,y index=2] {vary_noise/vary_noise_gt3_em.dat};

        \end{axis}
    \end{tikzpicture}%
    \begin{tikzpicture}
        \begin{axis}[
            scale only axis,
            height=2cm,
            width=\Width,
            ytick pos=left,
            ylabel={$d_{B}$},
            xlabel={$\sigma$},
            yticklabel style = {font=\footnotesize, color=red},
            xticklabel style = {font=\footnotesize},
            every y tick/.style={red},
            ylabel near ticks,
            xlabel near ticks,
            xmin=0,xmax=128,
            ymin=0,ymax=.4,
            ]
        \addplot[mark=none, red] table[x index=0,y index=1] {vary_noise/vary_noise_gt4.dat};
        \addplot[mark=none, red, dashed] table[x index=0,y index=1] {vary_noise/vary_noise_gt4_em.dat};

        \draw (rel axis cs:0,.88) node[right,text=white, fill=black,align=left] {GT4};
        \end{axis}
        
        \begin{axis}[
            scale only axis,
            height=2cm,
            width=\Width,
            axis y line*=right,
            axis x line=none,   
            ylabel={$\jac$},
            yticklabel style = {font=\footnotesize, color=blue},
            xticklabel style = {font=\footnotesize},
            every y tick/.style={blue},
            ylabel near ticks,
            xlabel near ticks,
            xmin=0,xmax=128,  
            ymin=0.2,ymax=1, 
            ]
        \addplot[mark=none, blue] table[x index=0,y index=2] {vary_noise/vary_noise_gt4.dat};
        \addplot[mark=none, blue, dashed] table[x index=0,y index=2] {vary_noise/vary_noise_gt4_em.dat};

        \end{axis}
    \end{tikzpicture}
    \vspace{-5pt}
    \caption{Varying the standard deviation of normal distributions on the images generated according to the GMM procedure under all available ground truth-masks. The estimation and segmentation results are shown in red and blue, respectively. Our method's results are plotted in solid lines, whereas the results arising from the EM-GMM algorithm are depicted in dashed lines. As in Figure \ref{fig:vary_slices}, the lines are the average results on 50 $300 \times 300$ images of $256$ colors. For segmentation, we set $\lambda = 1$ for both methods.}
    \label{fig:vary_noise}

\end{figure}
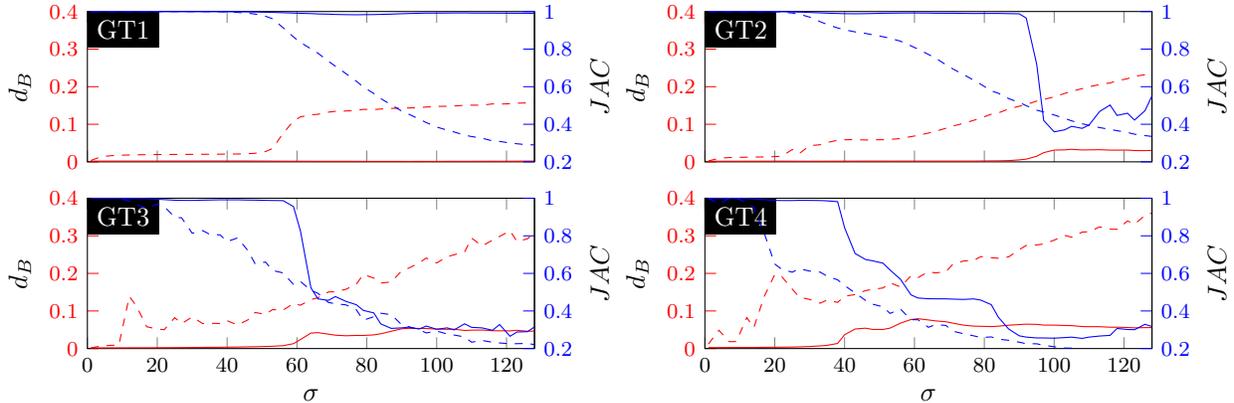

We start by evaluating TEAM and TEAMSEG under a controlled image generation setting, so we can assess several of its properties. Our image generation process consists of using the multi-region segmentation masks depicted in Figure \ref{fig:gts} to delineate the final image regions.  They were chosen following three criteria: (1) the regions they define have non-uniform sizes, (2) they present challenging features such as no-convex regions and fine structures, (3) they enable our assessment of model estimation and segmentation of images composed of up to 5 regions. More ground-truth masks and experiments illustrating other imaging settings could be approached, but we decided to restrict our focus to only these examples for experimental simplicity.

The pixel values in each segmentation region of each mask are generated according to two generative processes, named:
\begin{itemize}
    \item \textit{Gaussian Mixture Model} (GMM): Let $K$ be the number of regions in ground truth $G$. We pick $K$ maximally spaced values from 1 to $L$ (for example, $[1, 86, 171, 256]$ for $K =4$), and set the means of $K$ truncated normal distributions in $[1, L]$ with standard deviation $\sigma$ to those values. We assign a distribution to each region of $G$ and sample each of its pixels IID from that distribution.
    \item \textit{Random Appearance Models} (\Rand): Following \cite{rocha2022direct}, we sample $K$ random vectors in $\Delta_{L}$,  and assign each to a different region in $G$. We then sample the pixel values IID from that region according to that sampled distribution.
\end{itemize}

We chose these generative processes as (1) they allow for the easy creation of images with controlled sizes, number of colors, and model appearance overlap (for the case of GMM), all of which is analyzed in the next sections, (2) they generate images that follow the assumptions described in Section \ref{sec:assumptions}, being ideal for TEAM/TEAMSEG, and (3) for the case of GMM they produce realistic imaging data in some applications \cite{rodrigues2016sar}.

We assess our algorithms under two GMM settings, an easier and harder one, according to $\sigma$. Figure \ref{fig:example_synth_imgs} depicts samples of images synthesized using the process described above for each ground truth mask. 

For these experiments, Assumption \ref{as:independence} can be verified at $r = 1$, since we assume that the values for neighboring pixels are independently drawn from the distribution of the region they are in. Therefore, for images generated by both GMM and \Rand\ processes, we set $r = 1$, when computing $\alpha$, $\beta$, and $\gamma$ using TEAM.

\subsection{Varying the number of slices in $\gamma$}
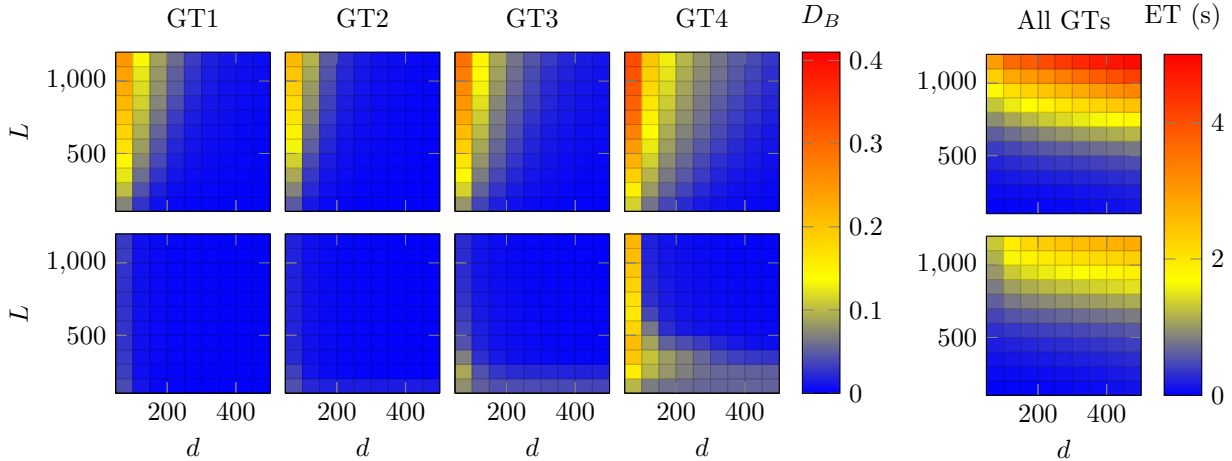
\begin{figure*}
    \begin{subfigure}[hb]{\linewidth}
\begin{tikzpicture}
        \begin{groupplot}[group style={group size= 4 by 2, vertical sep=0.3cm, horizontal sep=.2cm},
        view={0}{90},
        enlargelimits=false,
        point meta max=.41,
        point meta min=0,
        width=.22\linewidth,
        height = 3.7cm,
        ]
        \nextgroupplot[
            ylabel=$L$,
            xtick = {},
            xticklabels = {},
            title=GT1,
        ]
        \addplot3[surf, mesh/cols=10] table {vary_size_ncolors/vary_size_ncolors_dist_gt1_usegmm0.dat};   

        \nextgroupplot[
            ytick = {},
            yticklabels = {},
            xtick = {},
            xticklabels = {},
            title=GT2,
        ]
        \addplot3[surf, mesh/cols=10] table {vary_size_ncolors/vary_size_ncolors_dist_gt2_usegmm0.dat};

        \nextgroupplot[
            ytick = {},
            yticklabels = {},
            xtick = {},
            xticklabels = {},
            title=GT3,
        ]
        \addplot3[surf, mesh/cols=10] table {vary_size_ncolors/vary_size_ncolors_dist_gt3_usegmm0.dat};

        \nextgroupplot[
            colorbar right,
            colorbar style={
                title=$D_B$},
            every colorbar/.append style={height=
            2*\pgfkeysvalueof{/pgfplots/parent axis height}+ \pgfkeysvalueof{/pgfplots/group/vertical sep}},
            ytick = {},
            yticklabels = {},
            xtick = {},
            xticklabels = {},
            title=GT4,
        ]
        \addplot3[surf, mesh/cols=10] table {vary_size_ncolors/vary_size_ncolors_dist_gt4_usegmm0.dat};

        \nextgroupplot[
            ylabel=$L$,
            xlabel=$d$,
        ]
        \addplot3[surf, mesh/cols=10] table {vary_size_ncolors/vary_size_ncolors_dist_gt1_usegmm1.dat};

        \nextgroupplot[
            xlabel=$d$,
            ytick = {},
            yticklabels = {},
        ]
        \addplot3[surf, mesh/cols=10] table {vary_size_ncolors/vary_size_ncolors_dist_gt2_usegmm1.dat};

        \nextgroupplot[
            xlabel=$d$,
            ytick = {},
            yticklabels = {},
        ]
        \addplot3[surf, mesh/cols=10] table {vary_size_ncolors/vary_size_ncolors_dist_gt3_usegmm1.dat};

        \nextgroupplot[
            width=.22\linewidth,
            height = 3.7cm,
            xlabel=$d$,
            ytick = {},
            yticklabels = {},
        ]
        \addplot3[surf, mesh/cols=10] table {vary_size_ncolors/vary_size_ncolors_dist_gt4_usegmm1.dat};
          
        \end{groupplot}
\end{tikzpicture}
\end{subfigure}%
\hspace{-130pt}
\begin{subfigure}[hb]{.22\linewidth}
    \begin{tikzpicture}
        \begin{groupplot}[group style={group size= 1 by 2, vertical sep=0.3cm, horizontal sep=.2cm},
        view={0}{90},
        enlargelimits=false,
        point meta max=5,
        point meta min=0,
        width=\linewidth,
        height = 3.7cm,
        ]
        \nextgroupplot[
            colorbar,
            colorbar style={
                title=ET (s)},
            every colorbar/.append style={height=
            2*\pgfkeysvalueof{/pgfplots/parent axis height}+ \pgfkeysvalueof{/pgfplots/group/vertical sep}},
            xtick = {},
            xticklabels = {},
            title = {All GTs}
        ]
        \addplot3[surf, mesh/cols=10] table {vary_size_ncolors/vary_size_ncolors_times_average_usegmm0.dat};

        \nextgroupplot[
            xlabel=$d$,
        ]
        \addplot3[surf, mesh/cols=10] table {vary_size_ncolors/vary_size_ncolors_times_average_usegmm1.dat};

        \end{groupplot}
    \end{tikzpicture}
\end{subfigure}
    \caption{Varying image size and number of colors. For each heat map, we compute the average $D_B$ (per GT, on the left) and the average elapsed time (ET, averaged over all GTs, on the right) of 10 images generated of size $d \times d$ and with $L$ colors for $d \in [50, 100, \ldots 500]$ and $L \in [100, 200, \ldots, 1200]$. Images were generated using the \Rand\  approach on top rows, while GMM ($\sigma = 60$) was employed on the bottom ones.}\label{fig:colors_size}
\end{figure*}

As discussed in Section \ref{seg:dim_gamma}, our methods do not require that the computation of all slices in $\gamma$. Using fewer slices leads to a faster, but potentially worsened, estimation and to memory savings. Because of that, Figure \ref{fig:vary_slices} depicts the timing and the estimation performance of our method via the $D_B$ measure when varying the number of slices used during model estimation, according to what is described in Section \ref{seg:dim_gamma}. As expected, the estimation time linearly increases with $\ell$, the effective number of used slices from $\gamma$, and that result is consistent across different segmentation masks and for both image generation procedures. On the other hand, TEAM's estimation performance reaches a plateau after a certain value of $\ell$ (usually before a third of the total number of colors) and, from there, adding new slides does not improve the estimation. Using the data in Figure \ref{fig:vary_slices} and for experimental simplicity, we set $\ell = \lceil L/3\rceil$ for the next experiments in this experimental section. We show that this choice also leads to fast and performant estimators under different noise and imaging settings.   
\begin{table*}
\setlength{\fboxrule}{.5pt}
\setlength{\fboxsep}{0pt}
\centering
    \caption{Average $D_B$ (higher is better) and average $\jac$ (higher is better) between estimated and ground truth segmentations on 50 $300 \times 300$ synthetic images with $L = 256$ generated using different segmentation masks and image generation strategies. For each method, we assign the target parameter (TP) of each of its component algorithms. Some methodologies estimate both appearance models $\theta$, region proportions $w$, and segmentations $S$, while others only target some of those. The estimation elapsed times (ET) refer to the timing of the correspondent component algorithm. Our method uses $r = 1$ and $\lambda = 1$ in all experiments. Segmentation results within 0.005 of the best one overall are in bold.}
    \label{tab:quantitative_results_estimation}
    \begin{adjustbox}{width=\textwidth}
    \begin{threeparttable}[t]
\setlength{\tabcolsep}{4pt}

\begin{tabular}{cccccccccccccccccc}
&&&\multicolumn{12}{c}{\textbf{Image Setting}}& \\
\cmidrule(lr){4-15}
&&&
\multicolumn{4}{c}{GMM ($\sigma = 30$)} &
\multicolumn{4}{c}{GMM ($\sigma = 60$)} &
\multicolumn{4}{c}{\Rand} &\\
\cmidrule(lr){4-7} \cmidrule(lr){8-11} \cmidrule(lr){12-15} 
 \multicolumn{2}{c}{\textbf{Method}} & \textbf{TP} & GT1 & GT2  & GT3 & GT4 & GT1 & GT2  & GT3 & GT4  & GT1 & GT2  & GT3  & GT4 & \textbf{ET (s)}\\\midrule[1.5pt]

\multicolumn{2}{c}{\multirow{3}{*}[-.1cm]{TEAMSEG\tnote{*}}}
 &\shortstack[c]{$\theta$} &0.002 & 0.002 & 0.003 & 0.005 & 0.002 & 0.002 & 0.013 & 0.080 & 0.003 & 0.002 & 0.006 & 0.018 &\multirow{2}{*}{0.12} 
\\ 
    
 &&\shortstack[c]{$w$} &0.000 & 0.040 & 0.015 & 0.084 & 0.007 & 0.038 & 0.017 & 0.227 & 0.034 & 0.031 & 0.030 & 0.087 &
\\
\cmidrule(lr){3-16}
 &&\shortstack[c]{$S$} &    \textbf{\textbf{1.000}} & \textbf{0.995} & \textbf{0.988} & \textbf{0.989} & \textbf{0.995} & \textbf{0.992} & 0.978 & 0.472 & \textbf{0.988} & \textbf{0.988} & 0.984 & 0.906 &  1.72\tnote{$\dagger$}
\\
\midrule[1.5pt]
\multicolumn{2}{c}{\multirow{3}{*}[-.1cm]{EM-GMM\tnote{*}}}
&\shortstack[c]{$\theta$} &0.020 & 0.032 & 0.076 & 0.131 & 0.115 & 0.069 & 0.110 & 0.193 & 0.285 & 0.448 & 0.514 & 0.557 &\multirow{2}{*}{1.17} 
\\ 

&&\shortstack[c]{$w$} &
0.043 & 0.044 & 0.042 & 0.043 & 0.049 & 0.063 & 0.042 & 0.042 & 0.080 & 0.054 & 0.048 & 0.047 & 
\\
\cmidrule(lr){3-16}
&&\shortstack[c]{$S$} &
\textbf{\textbf{1.000}} & 0.971 & 0.827 & 0.618 & 0.849 & 0.809 & 0.568 & 0.351 & 0.333 & 0.210 & 0.155 & 0.120  &1.63

\\
 \midrule[1.5pt]
 \multicolumn{2}{c}{\multirow{2}{*}{ALGB\tnote{*}}}
 &\shortstack[c]{$\theta$} & 0.008 & -- & -- & -- & 0.006 & -- & -- & -- & 0.005 & -- & -- & -- &2.20 
\\
 &&\shortstack[c]{$S$}    & \textbf{1.000} & -- & -- & --  & \textbf{0.995} & -- & -- & --  & \textbf{0.986} & -- & -- & --  &0.25 
\\
 \midrule[1.5pt]
 \multicolumn{2}{c}{\multirow{2}{*}{SPEC\tnote{*}}}
 &\shortstack[c]{$\theta$} & 0.008 & -- & -- & -- & 0.006 & -- & -- & -- & 0.003 & -- & -- & -- &0.03 
\\
 &&\shortstack[c]{$S$}   & \textbf{1.000} & -- & -- & -- & \textbf{0.995} & -- & -- & -- & \textbf{0.987} & -- & -- & -- & 0.25

\\
\midrule[1.5pt]
\multirow{7}{*}[-.05cm]{{ALT}}
& \multirow{2}{*}{$\lambda = 1$}   & $\theta$  
 & 0.003 & 0.061 & 0.016 & 0.028 & 0.001 & 0.013 & 0.012 & 0.050 & 0.004 & 0.001 & 0.005 & 0.007 & \multirow{2}{*}{35.23} \\ 
&   & $S$  
 & \textbf{1.000} & 0.906 & 0.968 & 0.922 & \textbf{0.992} & 0.956 & 0.914 & 0.624 & 0.929 & 0.978 & 0.930 & 0.906  \\ 
\cmidrule(lr){3-16}
& \multirow{2}{*}{$\lambda = 2$}    & $\theta$    
 & 0.003 & 0.002 & 0.003 & 0.004 & 0.001 & 0.001 & 0.001 & 0.046 & 0.001 & 0.000 & 0.000 & 0.003 & \multirow{2}{*}{38.52} \\ 
&     & $S$    
  &\textbf{1.000} & \textbf{0.996} & \textbf{0.990} & \textbf{0.992} & \textbf{0.992} & \textbf{0.994} & \textbf{0.995} & 0.592 & \textbf{0.988} & \textbf{0.990} & \textbf{0.995} & \textbf{0.963} & \\ 
 \cmidrule(lr){3-16}

& \multirow{2}{*}{$\lambda = 5$}  & $\theta$    
 & 0.003 & 0.002 & 0.004 & 0.020 & 0.001 & 0.001 & 0.001 & 0.045 & 0.036 & 0.001 & 0.000 & 0.014 & \multirow{2}{*}{43.91}
 \\
&   & $S$    
& \textbf{1.000} & \textbf{0.996} & \textbf{0.990} & 0.971 & \textbf{0.992} & \textbf{0.994} & \textbf{0.992} & 0.597 & 0.849 & \textbf{0.990} & \textbf{0.991} & 0.875 &
 \\
 \midrule[1.5pt]
 \multicolumn{2}{c}{FSEG}& $S$&
\textbf{0.996} & \textbf{0.992} & \textbf{0.989} & 0.976 & \textbf{0.996} & 0.987 & 0.977 & \textbf{0.755} & 0.484 & 0.400 & 0.318 & 0.246 & 0.17 \\
 \midrule[1.5pt]
\multicolumn{2}{c}{ORTSEG}& $S$&
0.987 & 0.987 & \textbf{0.988} & 0.971 & 0.987 & 0.984 & 0.979 & 0.710 & 0.752 & 0.690 & 0.519 & 0.435 & 2.09 \\
\midrule[1.5pt]
\multicolumn{2}{c}{GRPNMF} & $S$&
0.949 & 0.329 & 0.135 & 0.136 & 0.769 & 0.264 & 0.146 & 0.120 & 0.345 & 0.188 & 0.110 & 0.068 & 15.08 \\
\bottomrule[1.5pt]
\end{tabular}
\begin{tablenotes}
    \item[*] Used the graph cut framework to perform segmentation using the estimated appearance models.
    \item[$\dagger$] Average time per ground-truth mask -- GT1: 0.25s, GT2: 1.31s, GT3: 2.52s, GT4: 3.46s.

\end{tablenotes}
\end{threeparttable}
    \end{adjustbox}
\end{table*}

\subsection{Varying noise intensity and appearance overlap}\label{sec:vary_noise}

In Figure \ref{fig:vary_noise}, we demonstrated our TEAM's estimation and TEAMSEG's segmentation performance under various values from $\sigma$ in the GMM image generations setting. In the same figure, we also compared those results to appearance model estimations from the EM-GMM algorithm, followed by the graph cut segmentation using these models. Our method is able to outperform the baseline in this setting for almost all tested values of $\sigma$ for both estimation and segmentation. This result is relevant as it demonstrates the capacity of the proposed algorithm to handle Gaussian data better than the application-specific EM method while still maintaining its generality. Furthermore, Figure \ref{fig:vary_noise} also illustrates the existence of a threshold $\sigma$ passed beyond which the distribution overlap is such that both algorithms cannot distinguish well between the appearance models and therefore do not accomplish good estimates. Notice, however, that TEAM's threshold is always further into the $\sigma$ scale than EM-GMM's.

These experiments elucidate one of the most impactful drawbacks of using appearance models that work directly on individual pixel colors for segmentation, as this model overlap greatly influences the final segmentation performance. From Figure \ref{fig:vary_noise}, we also notice that the estimation performed by TEAMSEG also worsens as the overlap becomes more noticeable.

\subsection{Varying number of colors and image size}

In Figure \ref{fig:colors_size}, we evaluate our estimator's performance on synthetic images of various sizes and present different amounts of colors, for both GMM and \Rand\ generation processes. We also keep track of the computational time spent on estimation for each image. From that data, we can draw some important conclusions. 

Firstly, these results bring evidence that TEAM is able to perform fast and efficient estimations on images of size greater than $200 \times 200$ pixels and $L < 500$, which is ideal in the case of high-resolution grayscale images. As expected, TEAM performs better estimations for images with more colors as the image size increases, which corresponds to the $\alpha$, $\beta$, and $\gamma$ being better estimated. On the other hand, however, the estimation time also increases with size (due to the estimation of $\gamma$, mainly) and with colors (primarily because of the SVD performed in Algorithm \ref{alg:estimation_models}). In any case, it remains less than 0.5 seconds for most tested image settings. Finally, it is also easy to see that the scarcity of data in small-sized images impacts negatively the estimation quality, which is also expected.

Secondly, the effect of the appearance overlap described in Section \ref{sec:vary_noise} is particularly explicit in the GMM data results depicted in Figure \ref{fig:colors_size}, especially for images generated from the GT4 mask. For low values of $L$, the overlap leads to comparably poor estimates, while this effect reverses when there are more colors and the overlap caused $\sigma = 60$ gradually becomes relatively small. Furthermore, the presence of more data, as the image size increases, improves the estimation, but does not completely solve the hindrance caused by the model overlap. 

\subsection{Performance Comparison on Synthetic Images}

In Table \ref{tab:quantitative_results_estimation}, we measure our model estimation and segmentation performance on a dataset of 50 synthetically generated images using both GMM (for two choices of $\sigma$) and \Rand\ generative processes. In it, we demonstrate that TEAMSEG is able to effectively estimate the models of and segment the images under both GMM and \Rand\ generative processes, except for GT4 with GMM ($\sigma = 60$) setting, which we already discussed in prior sections.

Compared to ALGB and SPEC, TEAMSEG produces similar quantitative results for the $K=2$, which is expected since the three methods consider the same image assumptions. On the other hand, our proposed method (1) is not constrained to the binary segmentation setting, (2) is able to reliably estimate $w$, and (3) is fast, which are some of the objectives of this paper. 

Overall, our method only underperforms in estimation and segmentation efficacy when compared to ALT with $\lambda=5$, which is also the slowest competitive algorithm. Furthermore, it is important to notice that ALT is prone to converge to local suboptimal segmentations \cite{rocha2022direct}, whereas TEAMSEG is not. In terms of segmentation, ORTSEG outperforms TEAMSEG in some scenarios, but it generally fails when under the \Rand setting. This is mainly due to its reliance on filtering techniques, which TEAMSEG is free from, being another advantage of our method.

Our segmentation results could be further improved by tuning the value of $\lambda$ for each image separately, which is reasonable as certain smoother ground truths are better detected with larger border regularization. We decided in this paper to keep it constant, as our focus is on the estimation piece of our method. 


\subsection{Varying $r$ and its effect on Texture Segmentation}
\begin{figure}
    \centering
    \input{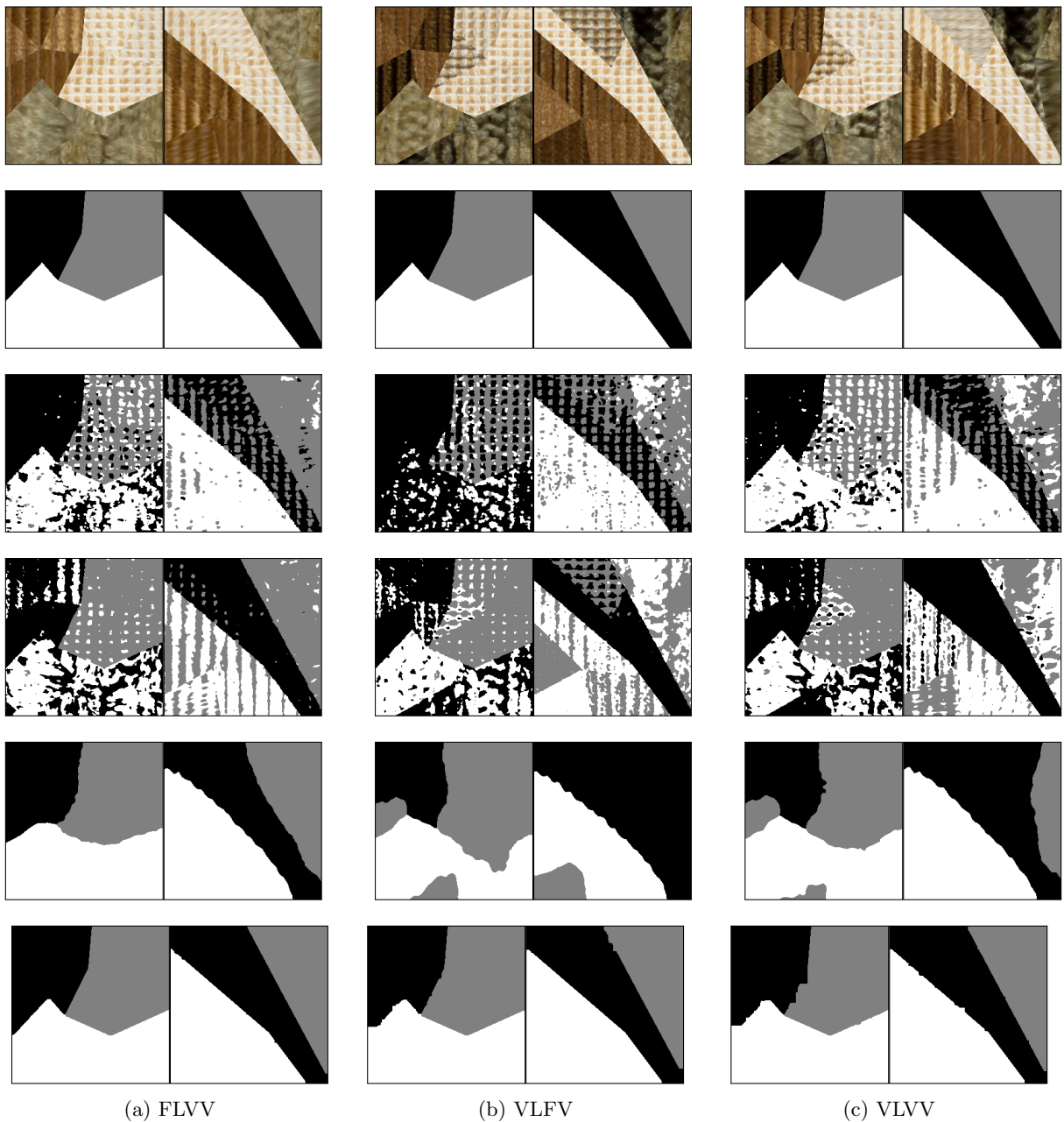}
    \caption{Selected segmentation results on the Prague Dataset. We show results from two images under the three available texture settings. The first row shows the original images, and the second the ground-truth masks. The third, fourth, and fifth rows depict the segmentations obtained by FSEG, ORTSEG, and GRPMNF, respectively. The last row shows TEAMSEG results for $r = 13$ and $\lambda = 5$. Image frames are not part of the original images.}
    \label{fig:prague}
\end{figure}

\begin{table}
\setlength{\fboxrule}{.5pt}
\setlength{\fboxsep}{0pt}
\centering
\caption{Average $\jac$ for BTF Texture Segmentation Data. For the proposed method, we set $\lambda = 5$ and $N = 256$ in all experiments. For TEAMSEG, the time columns comprise both estimation and segmentation runtimes. Segmentation results within 0.01 of the best one overall are in bold.}
  \label{tab:texture_results_estimation}
\begin{adjustbox}{max width=\textwidth}
\begin{tabular}{clcccc}
&&\multicolumn{3}{c}{\textbf{Dataset}}& \\
\cmidrule(lr){3-5}
\multicolumn{2}{c}{\textbf{Method}} &
FL/VV &
VL/FV &
VL/VV & \textbf{Time (s)}\\
\midrule[1pt]
\multirow{5}{*}{TEAMSEG}
 &$r = 1$   & 0.626 & 0.502 & 0.529 & 0.79 \\
 &$r = 8$   & 0.943 & 0.629 & 0.775 & 0.70 \\
 &$r = 13$  & \textbf{0.966} & \textbf{0.681} & \textbf{0.793} & 0.81 \\
 &$r = 18$  & \textbf{0.974} & \textbf{0.691} & 0.768 & 0.86 \\
 &$r = 26$  & 0.952 & 0.664 & \textbf{0.790} & 0.85 \\
\midrule[1pt]
\multicolumn{2}{c}{FSEG}
 &0.867  &  0.576 &   0.615 & 0.12
\\
\midrule[1pt]
\multicolumn{2}{c}{ORTSEG}
  &0.916    &0.678 &   0.771 & 1.33
\\
\midrule[1pt]
\multicolumn{2}{c}{GRPNMF}
& 0.618 & 0.483 & 0.556 & 3.71
\\
\bottomrule[1pt]
\end{tabular}
\end{adjustbox}
\end{table}

The prior experiments handled images where setting $r=1$ was a reasonable choice given the IID nature of the pixel data. In this section, we explore the effect of changing the value of $r$ so TEAMSEG is able to handle non-IID data or, specifically, textures. Following the results in \cite{rocha2022direct}, we use the fact that, for pairs $(x, y)$ of pixels distant by a large $r$ selected uniformly at random from $\Omega$, the values $I(x)$ and $I(y)$ are approximately independent. In this context, we are then able to make use of Assumption \ref{as:independence}. On the other hand, in multi-region images, the choice of a too large $r$ can worsen TEAM's estimation process for the reasons, as explained in Section \ref{sec:assumptions}.

To illustrate the impact of the value of $r$ in our method, we use the BTF Texture Segmentation Data from the Prague Texture Segmentation Datagenerator \cite{mikes2021texture}\footnote{Available at \url{https://mosaic.utia.cas.cz/}}. In it, we find three annotated segmentation datasets whose images were synthetically generated to have three regions of Bidirectional Texture Function (BTF) textures \cite{dana1999reflectance}, which are textured images whose pixel values can vary based on view and illumination spherical settings. Each region in each image is composed of five textures from a subset of BFT images that have either (1) fixed light and variable view (FLVV), (2) variable light and fixed view (VLFV), or variable light and variable view (VLVV); comprising therefore the three available datasets. Each dataset, here called FLVV, VLFV, and VLVV, is finally made of 90 $256 \times 256$ RGB images and $K=3$ ground truth regions. Some examples of these images can be found in Figure \ref{fig:prague}, where each column corresponds to a different dataset. These images were chosen such that they use the same ground-truth mask across datasets, illustrating their differences.

We chose these datasets to demonstrate our performance on textures for three reasons: (1) it is readily available for download, (2) there is less ambiguity in terms of how reliable the ground truth masks are, since each of the three subsets of textures in each dataset is visually different from the other two, reducing the regions' ambiguity, (3) the textures are realistic and challenging  enough for our purposes. 

Since these are colored images, we prepossessed the data before applying TEAMSEG, following the algorithm described in Section \ref{sec:color_imgs} and setting $N = 256$, for simplicity. In future work, this parameter can be tuned for better performance. In Table \ref{tab:texture_results_estimation}, we report the average segmentation performance of TEAMSEG for several values of $r$ alongside the tuned performances of FSEG, ORTSEG, and GRPNMF, which worked on the original data.  The other methods used the original images without prepossessing. Figure \ref{fig:prague} depicts the qualitative segmentation results from selected images. 

These results demonstrate the importance of the value of $r$ in TEAMSEG for images whose pixel values are not IID, as in the previous sections. Our method's performance greatly improves when we transition from $r=1$ to $r\in\{8, 13, 18\}$, indicating the satisfaction of Assumption \ref{as:independence} in this setting. That performance starts to decrease when we hit $r = 26$, demonstrating that Assumption \ref{as:short} may not be true in that context anymore. 

Overall, our method is also able to outperform all comparative methods, while maintaining a short model estimation and segmentation time. When contrasting the qualitative results from TEAMSEG and ORTSEG, the second best performing method, in Figure \ref{fig:prague}, we can conclude that TEAMSEG's success is partly due to ORTSEG's reliance on image filtering techniques that smooth the final segmentation borders. TEAMSEG is free from the use of image filters, which helps preserve sharp edges. A similar explanation can be the case of FSEG's and GRPNMF's failure in these datasets.

\subsection{Experiments on Real Images}
\begin{figure*}
\centering
\input{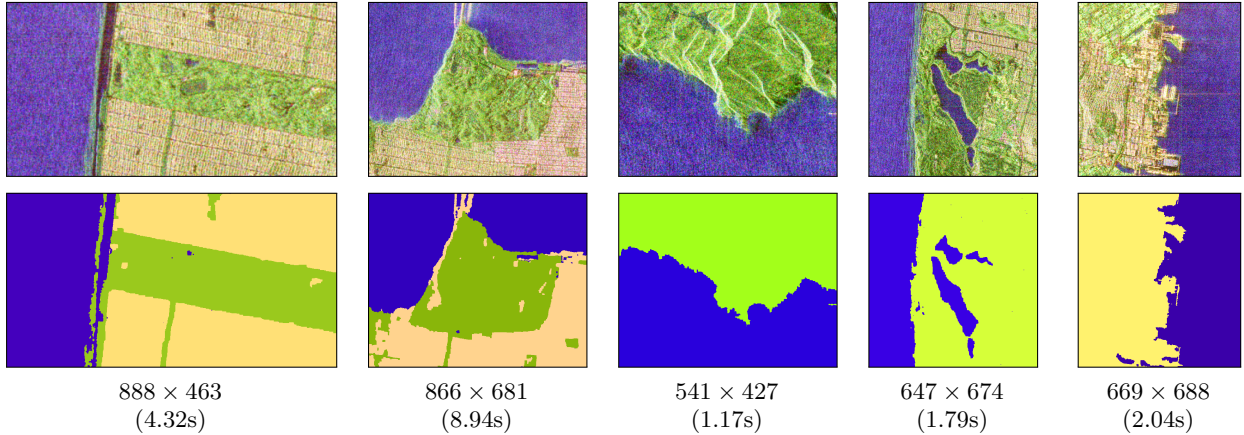}
\caption{TEAMSEG qualitative performance on PolSAR imagery. The top row displays the original cropped images, and the bottom one their retrospective segmentations obtained by TEAMSEG. Each image's size and the total model estimation + segmentation times are also shown. For all these results, we set $\lambda=2$ and $N = 256$ and the number of clusters is set differently for each image. Image frames are not part of the original images.}
\label{fig:real_sar}
\end{figure*}

\begin{figure*}
\centering
\input{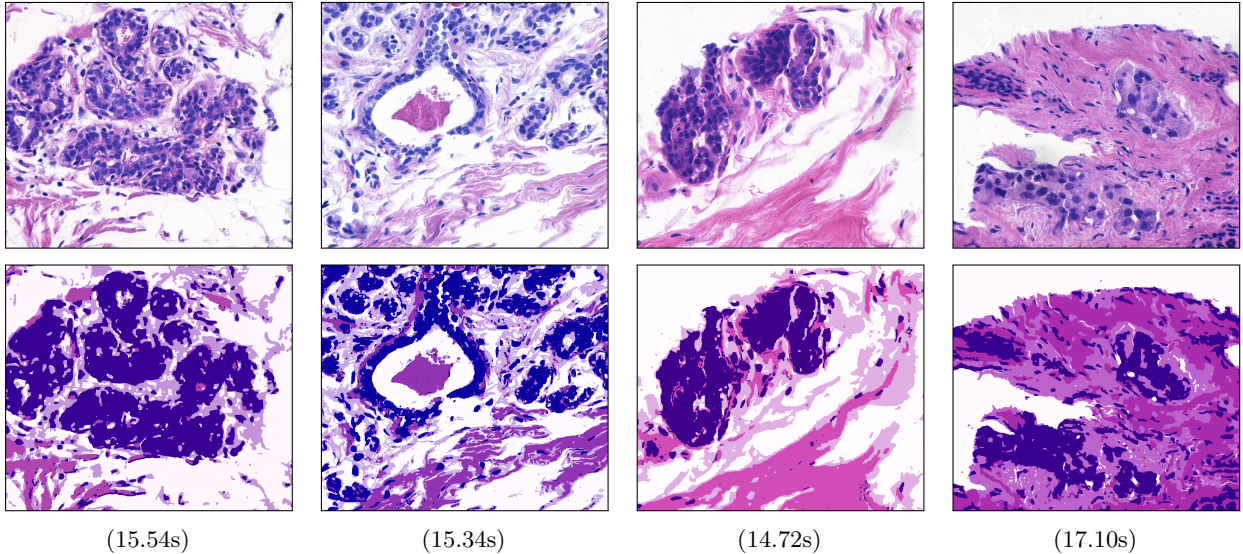}
\caption{TEAMSEG qualitative performance on histology images. Results are organized as in Figure \ref{fig:real_sar}.  All the images have all sizes $896 \times 768$ and we set $K = 4$, $\lambda=1$, and $N = 256$ for all experiments. Image frames are not part of the original images.}
\label{fig:real_hist}
\end{figure*}

Finally, our last set of results demonstrates TEAMSEG's performance on natural color images. Here, we keep $N = 256$, as in the previous section, and $r = 25$ and set a fixed $\lambda$ for each group of experiments.

Our first results are on Polarimetric Synthetic Aperture Radar (PolSAR) data, shown in Figure \ref{fig:real_sar}. These images are crops from an original high-resolution PolSAR image from the San Francisco Bay recorded from the satellite GAOFEN-3 \cite{xu2019polsf}\footnote{Original data available at \url{https://ietr-lab.univ-rennes1.fr/polsarpro-bio/san-francisco/}}. In them, we can immediately see the presence of three possible distinct regions: sea, vegetation, and urban areas. Based on the visual inspection of the crops to determine the number of regions present, we ran TEAMSEG and obtained the visually effective segmentation results shown in Figure \ref{fig:real_sar}. They demonstrate our algorithm's capacity to estimate appearances from an imagery type (PolSAR)  known to be hard to model, due to its inherent noise pattern \cite{khan2012single}. These results also elucidate how TEAMSEG is able to find precise, sharp boundaries between the regions.

The second set of selected images for experimentation are histopathology images used in breast cancer cell detection from the Center for Bio-Image Informatics at the University of California, Santa Barbara \cite{drelie2009biosegmentation}\footnote{Available at \url{https://bioimage.ucsb.edu/research/bio-segmentation}}. Here, we segment all these images into four regions: the background, cell nuclei, and two distinct intercellular regions (one of lighter and the other of darker color). We used these images to illustrate our method's capacity to segment very fine and small structures, such as individual nuclei separated from a large a nuclei mass, accurately. We also notice our method's ability to accurately detect the background regions, precisely delineating them from the other foreground areas. As a drawback, these results show how TEAMSEG is prone to blend nearby regions of similar appearance, as is the case of the multiple nuclei masses in the images. This phenomenon can potentially be avoided by tuning the parameter lambda according to each image.

\section{Conclusion and Future Work}
In this work, we explored the problem of estimation of appearance models directly from the image. The solution for this problem is relevant to the problem of image segmentation, as appearance models are crucial for many of its graph-based solvers. In the recent past, some solutions for the binary region case were proposed, but they were not able to be easily generalized for the multi-region case, nor did they give estimates for the region propositions directly. In this work, we propose an estimation algorithm, called TEAM, that is able to tackle these drawbacks by using third-order statistics from the image. When coupled with the graph-cut algorithm for segmentation, our overall method TEAMSEG is able to outperform many competitive algorithms in most experimental tasks involving IID data while maintaining competitive model estimation and segmentation runtimes. Furthermore, the proposed method can also achieve satisfactory texture segmentation results on a challenging dataset. This is relevant as TEAMSEG does not require the usage of filtering techniques in its algorithm, which makes it faster and also more apt to handle image regions with sharp corners and small artifacts. Finally, our methodology is also successful in segmenting PolSAR and histology images, which shows its potential to be applied in other remote sensing and biomedical imaging settings. 

Based on some applications of the work in  \cite{anandkumar2014tensor, ruffini2017hierarchical} to the problem of Topic Modeling from Natural Language Processing, future work will focus on bridging this task with that of Model-based Image Segmentation. An effort in that direction could contribute with algorithms to both fields, as they are both fertile ground for theoretical and practical studies. On the implementation side, our work lacks further experimentation on the effect of the color-space partitioning for RGB images. It would also be improved with different solutions for the memory requirement issue it imposes when dealing with such data. Future work will also aim at furthering these studies. 

\bibliographystyle{IEEEtran} 
\bibliography{refs} 

\end{document}